\documentclass[10pt,twocolumn,letterpaper]{article}

\usepackage{iccv}
\usepackage{times}
\usepackage{epsfig}
\usepackage{graphicx}
\usepackage{amsmath}
\usepackage{amssymb}
\usepackage{float}

\usepackage{algorithm}
\usepackage{algorithmic}
\usepackage{wrapfig}

\usepackage{array}
\usepackage{multirow}
\newcolumntype{M}[1]{>{\centering\arraybackslash}m{#1}}
\newcolumntype{P}[1]{>{\centering\arraybackslash}p{#1}}

\usepackage[pagebackref=true,breaklinks=true,letterpaper=true,colorlinks,bookmarks=false]{hyperref}

\newcommand{\x}{{\bf x}}
\newcommand{\w}{{\bf w}}
\newcommand{\W}{{\bf \mathsf{W}}}

\newtheorem{theor}{Theorem}

\iccvfinalcopy

\ificcvfinal
\begin{document}

%%%%%%%%% TITLE
\title{Introspective Generative Modeling: Decide Discriminatively}

\author{Justin Lazarow $^*$ \\
Dept. of CSE\\
UCSD\\
{\tt\small jlazarow@ucsd.edu}
\and
Long Jin$^*$\\
Dept. of CSE\\
UCSD\\
{\tt\small lojin@ucsd.edu}
\and
Zhuowen Tu\\
Dept. of CogSci\\
UCSD\\
{\tt\small ztu@ucsd.edu}
}

\maketitle

\begin{abstract}
We study unsupervised learning by developing introspective generative modeling (IGM) that attains a generator using progressively learned deep convolutional neural networks.  The generator is itself a discriminator, capable of introspection: being able to self-evaluate the difference between its generated samples and the given training data. When followed by repeated discriminative learning, desirable properties of modern discriminative classifiers are directly inherited by the generator. IGM learns a cascade of CNN classifiers using a synthesis-by-classification algorithm. In the experiments, we observe encouraging results on a number of applications including texture modeling, artistic style transferring, face modeling, and semi-supervised learning. \footnote{$^*$ equal contribution.}
\end{abstract} 

%\vspace{-2mm}
\section{Introduction}
\vspace{-2mm}

Supervised learning techniques have made substantial impact on tasks that can be formulated as a classification/regression problem; some well-known classifiers include SVM \cite{vapnik1995nature}, boosting \cite{AdaBoost}, random forests \cite{breiman2001random}, and convolutional neural networks \cite{CNN}. Unsupervised learning, where no task-specific labeling/feedback is provided on top of the input data, still remains one of the most difficult problems in machine learning but holds a bright future since a large number of tasks have little to no supervision.

Popular unsupervised learning methods include mixture models \cite{duda2000pattern}, principal component analysis (PCA) \cite{jolliffe2002principal}, 
spectral clustering \cite{shi2000normalized}, topic modeling \cite{blei2003latent}, and autoencoders \cite{bengio2007scaling, baldi2012autoencoders}.
In a nutshell, unsupervised learning techniques are mostly guided by the minimum description length principle (MDL) \cite{rissanen1978modeling} to best reconstruct the data whereas supervised learning methods are primarily driven by minimizing error metrics to best fit the input labeling. Unsupervised learning models are often generative and supervised classifiers are often discriminative; generative model learning has been traditionally considered to be a much harder task than discriminative learning \cite{friedman2001elements}, due to its intrinsic learning complexity, as well many assumptions and simplifications made about the underlying models.

\begin{figure}[!htp]
\vspace{-2mm}
\begin{center}
\begin{tabular} {c}
\includegraphics[width=0.46\textwidth]{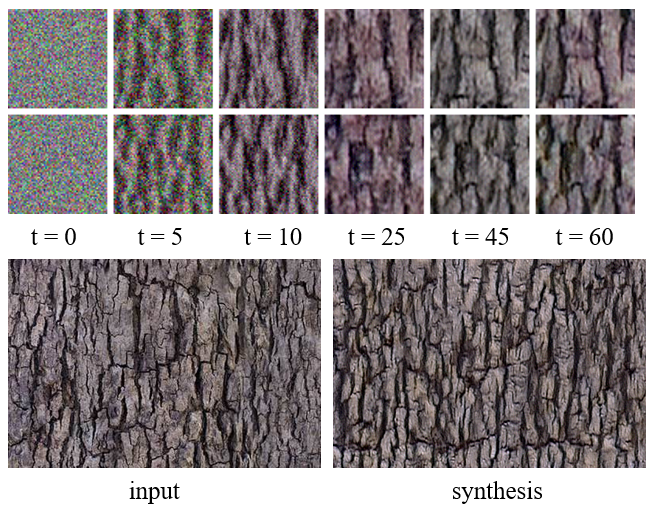}
\end{tabular}
\end{center}
\vspace{-5mm}
\caption{\footnotesize The first row shows the development of two $64 \times 64$ pseudo-negative samples (patches) over the course of the training process on the ``tree bark'' texture at selected rounds. We can see the initial ``scaffold'' created and then refined by the network in later rounds. The input ``tree bark'' texture and a synthesized image by our IGM algorithm are shown in the second row.}
\label{fig:illustration}
\vspace{-2mm}
\end{figure}

Generative and discriminative models have traditionally been considered distinct and complementary to each other. In the past, connections have been built to combine the two families \cite{friedman2001elements,liang2008asymptotic,tu2008brain,jebara2012machine}. In the presence of supervised information with a large amount of data, a discriminative classifier \cite{krizhevsky2012imagenet} exhibits superior capability in making robust classification by learning rich and informative representations; unsupervised generative models do not require supervision but at a price of relying on assumptions that are often too ideal in dealing with problems of real-world complexity. Attempts have previously been made to learn generative models directly using discriminative classifiers for density estimation \cite{welling2002self} and image modeling \cite{tu2007learning}. There is also a wave of recent development in generative adversarial networks (GAN) \cite{goodfellow2014generative,radford2015unsupervised,salimans2016improved,arjovsky2017wasserstein} in which a discriminator helps a generator try not to be fooled by ``fake'' samples. We will discuss in detail the relations and connections of our IGM with these existing literature in the next section.  

In \cite{welling2002self}, a self supervised boosting algorithm was proposed to train a boosting algorithm by sequentially adding features as weak classifiers on additionally self-generated negative samples; the generative discriminative modeling work (GDL) in \cite{tu2007learning} generalizes the concept that a generative model can be successfully modeled by learning a sequence of discriminative classifiers via self-generated pseudo-negatives.

Inspired by the prior work on generative modeling \cite{zhu1997minimax,welling2002self,tu2007learning} and development of convolutional neural networks \cite{CNN,krizhevsky2012imagenet,gatys2015neural}, we develop an image modeling algorithm, introspective generative modeling (IGM) that is simultaneously a generator and a discriminator, consisting of two critical stages during training: (1) a pseudo-negative sampling stage (synthesis) for self-generation, (2) and a CNN classifier learning stage (classification) for self-evaluation and model updating.  There are a number of interesting properties about IGM that are worth highlighting:
%\vspace{-1mm}
{\small 
%\vspace{-1.0mm}
\begin{itemize}
\item {\bf CNN classifier as generator}: No special condition on CNN architecture is needed in IGM and many existing CNN classifiers can be directly made into generators, if trained properly.
%\vspace{-1.0mm}
\item {\bf End-to-end self-evaluation and learning}: Perform end-to-end ``introspective learning'' to self-classify between synthesized samples (pseudo-negatives)  and the training data, followed by direct discriminative learning, to approach the target distribution.
%\vspace{-1.0mm} 
\item {\bf Integrated unsupervised/supervised learning}: Unsupervised and supervised learning can be carried out under similar pipelines, differing in the absence or presence of initial negative samples.
%\vspace{-1.0mm}
\item {\bf All backpropagation}: Our synthesis-by-classification algorithm performs efficient training using backpropagation in both stages: the sampling stage for the input images and the classification stage for the CNN parameters.
%\vspace{-1.0mm}
\item {\bf Model-based anysize-image-generation}: Since our generative modeling models the input image, we are able to train on images of a size and generate an image of a larger size while maintaining the coherence for the entire image.
%\vspace{-1.0mm}
\item {\bf Agnosticity to various vision applications}: Due to its intrinsic modeling power being  at the same time generative and discriminative, IGM can be adopted to many applications; under the same pipeline, we show a number of vision tasks in the experiments, including texture modeling, artistic style transference, face modeling, and semi-supervised image classification in this paper. Supervised classification cases of an algorithm in the same introspective learning family can been seen in \cite{jing2017Introspective}.
%\vspace{-1.0mm}
\end{itemize}
\vspace{-2mm}
}

\vspace{-2mm}
\section{Significance and related work}
\vspace{-2mm}

Our introspective generative modeling (IGM) algorithm has connections to many existing approaches including the MinMax entropy work for texture modeling \cite{zhu1997minimax}, the hybrid modeling work \cite{friedman2001elements}, and the self-supervised boosting algorithm \cite{welling2002self}. It builds on top of convolutional neural networks \cite{CNN} and we are particularly inspired by two lines of prior algorithms: the generative modeling via discriminative approach method (GDL) \cite{tu2007learning}, and 
the DeepDream code \cite{mordvintsev2016deepdream} and the neural artistic style work \cite{gatys2015neural}. The general pipeline of IGM is similar to that of GDL \cite{tu2007learning}, with the boosting algorithm used in \cite{tu2007learning} is replaced by a CNN in IGM. More importantly, the work of \cite{mordvintsev2016deepdream,gatys2015neural} motives us to significantly improve the time-consuming sampling process in \cite{tu2007learning} by an efficient SGD process via backpropagation (the reason for us to say ``all backpropagation''). Next, we review some existing generative image modeling work, followed by detailed discussions about the two most related algorithms: GDL \cite{tu2007learning} and the recent development of generative adversarial networks (GAN)  \cite{goodfellow2014generative}.

The history of generative modeling on image or non-image domains is extremely rich, including the general image pattern theory \cite{grenander1993general}, deformable models \cite{yuille1992feature}, inducing features \cite{della1997inducing}, wake-sleep \cite{hinton1995wake}, the MiniMax entropy theory \cite{zhu1997minimax}, the field of experts \cite{roth2005fields}, Bayesian models \cite{yuille2006vision}, and deep belief nets \cite{Hinton06}. Each of these pioneering works points to some promising direction to unsupervised generative modeling. However the modeling power of these existing frameworks is still somewhat limited in computational and/or representational aspects. In addition, not too many of them sufficiently explore the power of discriminative modeling. Recent works that adopt convolutional neural networks for generative modeling \cite{xie2016theory} either use CNNs as a feature extractor or create separate paths \cite{xie2016cooperative,ulyanov2016texture}. The neural artistic transferring work \cite{gatys2015neural} has demonstrated impressive results on the image transferring and texture synthesis tasks but it is focused \cite{gatys2015neural} on a careful study of channels attributed to artistic texture patterns, instead of aiming to build a generic image modeling framework. The self-supervised boosting work \cite{welling2002self} sequentially learns weak classifiers under boosting \cite{AdaBoost} for density estimation, but its modeling power was not adequately demonstrated.

\vspace{-2mm}
\subsection*{Relationship with GDL \cite{tu2007learning}}
\vspace{-2mm}

The generative via discriminative learning framework (GDL) \cite{tu2007learning}
learns a generator through a sequence of boosting classifiers \cite{AdaBoost} using repeatedly self-generated samples, called {\bf pseudo-negatives}, to approach the target distribution.
Our IGM algorithm takes inspiration from GDL, but we also observe a number of limitations in GDL that will be overcome by IGM: GDL uses manually specified feature types (histograms and Haar filters), which are fairly limited by today's standard; the sampling process in GDL, based on Markov chain Monte Carlo (MCMC), is a big computational bottleneck; the experimental results for image modeling and classification were not satisfactory. To summarize, the main differences between GDL and IGM include: 
{\small
\vspace{-1mm}
\begin{itemize}
\item The adoption of convolutional networks in IGM results in a significant boost to feature learning.
\item Introducing backpropagation to the synthesis/sampling process in IGM makes a fundamental improvement to the sampling process in GDL that is otherwise slow and impractical.
\item An alternative algorithm, namely IGM-single (see Fig. \ref{fig:IGM_single}), is additionally proposed to maintain a single classifier for IGM.
\item Higher quality results for image modeling are demonstrated in IGM.
\end{itemize}
}
\vspace{-2mm}
\subsection*{Comparison with GAN \cite{goodfellow2014generative} }
\vspace{-2mm}
The recent development of generative adversarial neural networks \cite{goodfellow2014generative} is very interesting and also highly related to IGM. We summarize the key differences between IGM and GAN. Other recent algorithms alongside GAN \cite{goodfellow2014generative,radford2015unsupervised,zhao2016energy,brock2016neural,tolstikhin2017adagan} share similar properties with it.

{\small
\begin{itemize}
\item {\em Unified generator/discriminator vs. separate generator and discriminator}.  IGM maintains a single model that is simultaneously a generator and a discriminator. The IGM generator therefore has self-awareness --- being able to self-evaluate the difference between its generated samples (called pseudo-negatives) w.r.t. the training data, followed by a direct CNN classifier training. GAN instead creates two convolutional networks, a generator and a discriminator.

\item {\em Training}. Due to the internal competition between the generator and the discriminator, GAN is known to be hard to train \cite{arjovsky2017wasserstein}. IGM carries out a straightforward use of backpropagation in both the sampling and the classifier training stage, making the learning process direct. For example, for the textures shown in the experiments Fig. \ref{fig:texture_big} and Fig. \ref{fig:texture_small}, all results by IGM are obtained under the identical setting without hyper-parameter tuning.
\item {\em Modeling}. The generator in GAN is a mapping from the features to the images. IGM directly models the underlying statistics of an image  with an efficient sampling/inference process, which makes IGM flexible. For example, we are able to conduct model-based-anysize-generation in the texture modeling task by directly maintaining the underlying statistics of the entire image.

\item {\em Speed}. GAN performs a forward pass to reconstruct an image, which is generally faster than IGM where synthesis is carried out using backpropagation. IGM is still practically feasible since it takes only about $1-2$ seconds to synthesize an image of size $64 \times 64$ and around $30$ seconds to synthesize a texture image of size $320 \times 200$ excluding the time to load the models.
\item {\em Model size}. Since a cascade of CNN classifiers ($60-200$) are included in a single IGM model, IGM has a much larger model complexity than GAN. This is an advantage of GAN over IGM. Our alternative IGM-single model maintains a single CNN classifier but its generative power is worse than those of IGM and GAN.
\end{itemize}
}

\vspace{-2mm}
\subsection*{Relationship with ICL \cite{jing2017Introspective}}
\vspace{-2mm}

The Introspective Classifier Learning work (ICL) \cite{jing2017Introspective}  is a sister paper to IGM, with ICL focusing on the discriminator side emphasizing its classification power. {\bf a).} IGM focuses on the generator side studying its image construction capability. {\bf b).} IGM consists of a sequence of cascading classifiers, whereas ICL is only composed of a single classifier. Essentially, ICL is similar to IGM-single, with a small difference in the absence/presence of given negative samples. The generative modeling aspect of ICL/IGM-single is not as competitive as IGM though. {\bf c)}. In ICL, a formulation for training a softmax multi-class classification was proposed, which is not in IGM. {\bf d).} In addition, ICL focuses on single image patch, whereas IGM is able to model/synthesize an arbitrary size of image. A number of important image modeling tasks, including texture modeling,  style transferring, face modeling, and semi-supervised learning are demonstrated here, which are not covered in ICL \cite{jing2017Introspective}.

\vspace{-2mm}
\section{Method}
\vspace{-2mm}

We describe below the introspective generative modeling (IGM) algorithm. We discuss the main formulation first, which bears some level of similarity to GDL \cite{tu2007learning}. However, with the replacement of the boosting algorithm \cite{AdaBoost} by convolutional neural networks \cite{CNN}, IGM demonstrates  significant improvement over GDL in terms of both modeling and computational power. The enhanced modeling power comes mainly from CNNs due to its end-to-end learning with automatic feature learning and tuning when backpropagating on the network parameters; enhanced computational power also largely from CNNs due to a natural implementation of sampling by backpropagating on the input image.
GDL is similar to IGM (see Fig. \ref{fig:IGM_cascade}), but IGM-single (see Fig. \ref{fig:IGM_single}) maintains a single CNN as opposed to having a sequence of classifiers in both GDL and IGM. We motivate the formulation of IGM from the Bayes theory, similar to GDL.

\vspace{-2mm}
\subsection{Formulation}
\vspace{-2mm}

\begin{figure*}[!htp]
\begin{center}
\begin{tabular} {c}
\hspace{-5mm} \includegraphics[width=1.0\textwidth]{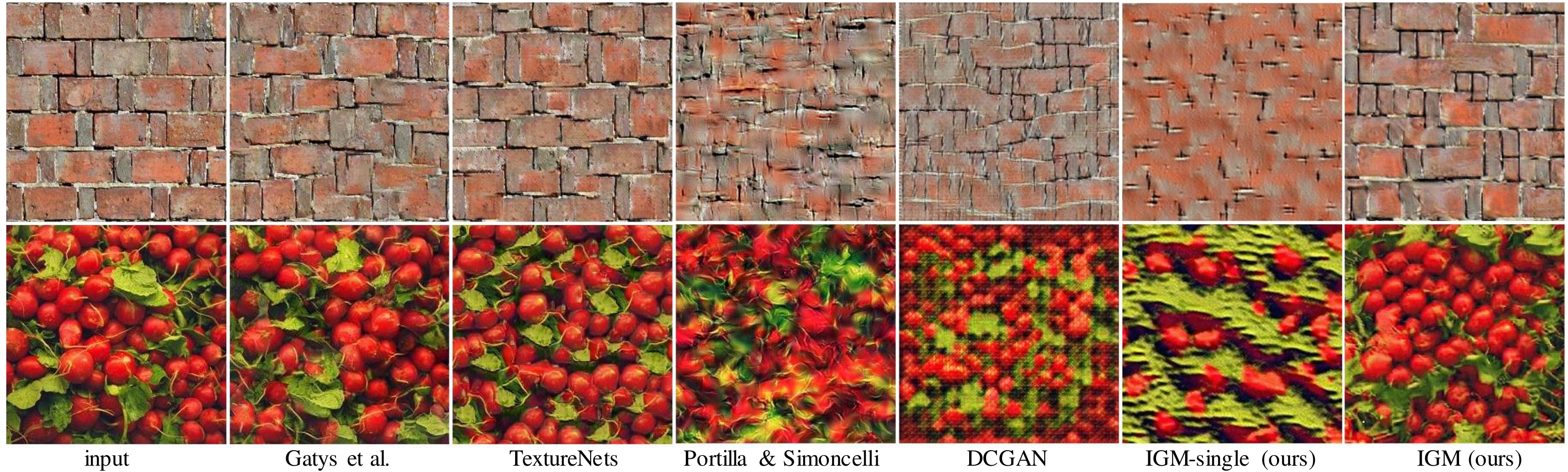}
\end{tabular}
\end{center}
\vspace{-3mm}
\caption{\footnotesize Texture synthesis algorithm comparison. Gatys et al. \cite{gatys2015neural}, Texture Nets \cite{ulyanov2016texture}, Portilla $\&$ Simoncelli \cite{portilla2000parametric}, and DCGAN \cite{radford2015unsupervised} results are from \cite{ulyanov2016texture}.}
\label{fig:texture_big}
\vspace{-2mm}
\end{figure*}

We start the discussion by borrowing notation from \cite{tu2007learning}.
Suppose we are given a set of training images (patches): $S=\{\x_i, i=1..n\}$.
We focus on patch-based input first and let $\x \in {\cal R}^m$ be a data sample (an image patch of size say $m = 64 \times 64$). We adopt the pseudo-negative concept defined in \cite{tu2007learning} and define class labels $y \in \{-1, +1 \}$, indicating $\x$ being a negative or a positive sample. Here we assume the positive samples with label $y=+1$ are the patterns/targets we want to study. A generative model computes for $p(y,\x)=p(\x|y) p(y)$, which captures the underlying
generation process of $\x$ for class $y$. A discriminative classifier instead computes $p(y|\x)$.
Under the Bayes rule, similar to the motivation in \cite{tu2007learning}:
\vspace{-2mm}
\begin{equation}
  p(\x|y=+1) = \frac{p(y=+1|\x) p(y=-1)}{p(y=-1|\x) p(y=+1)} p(\x|y=-1),
\end{equation}
which can be further simplified when assuming equal priors $p(y=+1)= p(y=-1)$:
\begin{equation}
  p(\x|y=+1) = \frac{p(y=+1|\x)}{1-p(y=+1|\x)} p(\x|y=-1).
\label{eq:gen1}
\end{equation}
\vspace{-3mm}

Based on Eq. (\ref{eq:gen1}), a generative model for the positive samples (patterns of interest) $p(\x|y=+1)$ can be fully represented by a generative model for the negatives $p(\x|y=-1)$ and a discriminative classifier $p(y=+1|\x)$, if both $p(\x|y=-1)$ and $p(y=+1|\x)$ can be accurately obtained/learned. However, this seemingly intriguing property is, in a way, a chicken-and-egg problem. To faithfully learn the positive patterns $p(\x|y=+1)$, we need to have a representative $p(\x|y=-1)$, which is equally difficult, if not more. For clarity, we now use $p^-(\x)$ to represent $p(\x|y=-1)$. In the GDL algorithm \cite{tu2007learning}, a solution was given to learning $p(\x|y=+1)$ by using an iterative process starting from an initial reference distribution of the negatives $p_0^-(\x)$, e.g. a Gaussian distribution $U(\x)$ on the entire space of $\x \in {\cal R}^m$:
\[
\vspace{-2mm}
 p_0^-(\x) = U(\x)
\]
\begin{equation}
 p_{t}^-(\x) =  \frac{1}{Z_{t}} \frac{q_{t} (y=+1|\x)}{q_{t}(y=-1|\x)} \cdot p_{t-1}^-(\x), \; t=1..T
\label{eq:gdl1}
\end{equation}
where $Z_t=\int \frac{q_t (y=+1|\x)}{q_t(y=-1|\x)} p^-_{t-1}(\x) d\x$.
Our hope is to gradually learn $p_t^-(\x)$ by following this iterative process of Eq. \ref{eq:gdl1}:
\vspace{-2mm}
\begin{equation}
p_t^-(\x) \stackrel{t=\infty}{\rightarrow} p(\x|y=+1),
\label{eq:pt_def}
\end{equation}
such that the samples drawn $\x \sim p_t^-(\x)$
become indistinguishable from the given training samples.
The samples drawn from $\x \sim p_t^-(\x)$ are called {\bf pseudo-negatives}, following a definition in \cite{tu2007learning}. Next, we present the realization of Eq. \ref{eq:gdl1}, namely  IGM (consisting of a sequence of CNN classifiers and see Fig. \ref{fig:IGM_cascade}) and additionally IGM-single (maintaining a single CNN classifier and see Fig. \ref{fig:IGM_single}).

\vspace{-2mm}
\subsection{IGM Training}
\vspace{-2mm}

\begin{figure}[!htb]
\vspace{-3mm}
\begin{center}
\begin{tabular} {c}
\hspace{-5mm} \includegraphics[width=0.5\textwidth]{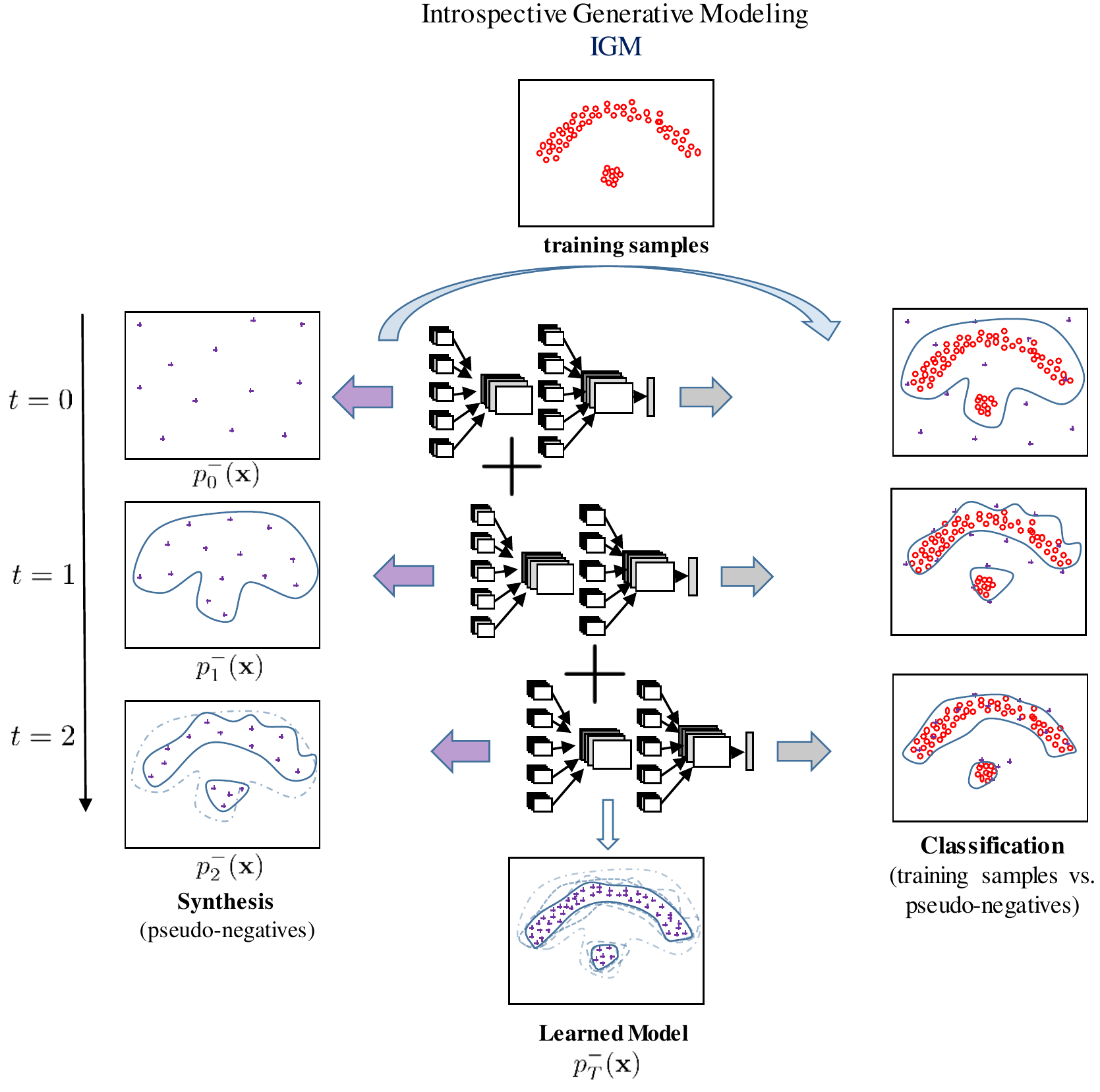} \\
\end{tabular}
\end{center}
\vspace{-3mm}
\caption{\footnotesize Schematic illustration of the pipeline of IGM. The top figure shows the input training samples shown in red circles. The bottom figure shows the pseudo-negative samples drawn by the learned final model. The left panel displays pseudo-negative samples drawn at each time stamp $t$. The right panel shows the classification by the CNN on the training samples and pseudo-negatives at each time stamp $t$.}
\label{fig:IGM_cascade}
\vspace{-2mm}
\end{figure}

\begin{algorithm}[!htp]
\caption{\small Outline of the IGM algorithm.}
\begin{algorithmic}
{\small
  \STATE {\bf Input: \hspace{2mm}} Given a set of training data $S_+=\{(\x_i,y_i=+1),i=1..n\}$ with $\x \in \Re^m$. \\
  \STATE {\bf  Initialization}: obtain an initial distribution e.g. Gaussian for the pseudo-negative samples: $p_0^-(\x) = U(\x)$. Create $S_{-}^0 = \{(\x_i,-1),i=1,...,l \}$ with $\x_i \sim p_0^-(\x)$\\
  \STATE {\bf  For} t=1..T \\
	\STATE {\bf 1.} Classification-step: Train CNN classifier $C^{t}$ on $S_+ \cup S_{-}^{t-1}$, resulting in $q_{t}(y=+1|\x)$.\\
	\STATE {\bf 2.} Update the model: $p_{t}^-(\x) =  \frac{1}{Z_t} \frac{q_t (y=+1|\x)}{q_t(y=-1|\x)} p_{t-1}^-(\x)$.\\
  \STATE {\bf 3.} Synthesis-step: sample $l$ pseudo-negative samples $\x_i \sim p_t^-(\x), i=1,...,l$ from the current model $p_t^-(\x)$ using a variational sampling procedure (backpropagation on the input) to obtain $S_{-}^{t} =  \{(\x_i,-1), i=1,...,l \}$. \\
  \STATE {\bf 4.} $t \leftarrow t+1$ and go back to step 1 until convergence (e.g. indistinguishable to the given training samples).
	\STATE {\bf End}
}
\end{algorithmic}
\label{al:IGM-cascade}
\end{algorithm}
\vspace{-1mm}
Next, we present our introspective generative modeling algorithm using a sequence of classifiers, called IGM. The given (unlabeled) training set is defined as $S=\{\x_i, i=1..n\}$, which is turned into $S_{+}=\{(\x_i,y_i=+1), i=1..n\}$ within the discriminative setting. We start from an initial pseudo-negative set
\[
   S_{-}^0 = \{(\x_i,-1),i=1,...,l \}
\]
where $\x_i \sim  p_0^-(\x)=U(\x)$ which is a Gaussian distribution.
A working set for $t=1..T$
\vspace{-2mm}
\[
S_{-}^{t-1} = \{(\x_i,-1),i=1,...,l \}.
\]
then includes the pseudo-negative samples self-generated from each round.
$l$ indicates the number of pseudo-negatives generated at each round.
We carry out learning with $t=1...T$ to iteratively obtain
\vspace{-2mm}
\begin{equation}
q_{t}(y=+1|\x), \;\; q_{t}(y=-1|\x)
\end{equation}
by updating classifier $C^t$ on $S_{+} \cup S_{-}^{t-1}$.
The reason for using $q$ is because it is an approximation to the true $p$ due to limited samples drawn in $\Re^m$. At each time $t$, we then compute
\vspace{-2mm}
\begin{equation}
 p_{t}^-(\x) =  \frac{1}{Z_{t}} \frac{q_{t} (y=+1|\x)}{q_{t}(y=-1|\x)} p_{t-1}^-(\x),
\label{eq:ICL}
\end{equation}
\vspace{-2mm}
where $Z_t=\int \frac{q_t (y=+1|\x)}{q_t(y=-1|\x)} p^-_{t-1}(\x) d\x$. We draw new samples
\[
\x_i \sim p_t^-(\x)
\]
\vspace{-2mm}
to have the pseudo-negative set:
\begin{equation}
{\small
 S_{-}^t = \{(\x_i,-1), i=1,...,l \}.
}
\end{equation}
Algorithm \ref{al:IGM-cascade} describes the learning process.
 The pipeline of IGM is shown in Fig. \ref{fig:IGM_cascade}, which consists of (1) a synthesis step and (2) a classification step. A sequence of CNN
classifiers is progressively learned. With the pseudo-negatives being gradually generated, the classification boundary gets tightened and approaches the target distribution. 

\vspace{-2mm}
\subsubsection{Classification-step}
\vspace{-2mm}
The classification-step can be viewed as training a normal classifier on the training set $S_+ \cup S_{-}^t$ where $S_{+}=\{(\x_i,y_i=+1),i=1..n\}$. $S_{-}^t=\{(\x_i,-1), i=1,...,l \}$ for $t\ge 1$. We use a CNN as our base classifier. When training a classifier $C^t$ on $S_+ \cup S_{-}^t$, we denote the parameters to be learned in $C^t$ by a high-dimensional vector $\W_t=(\w_t^{(0)}, \w_t^{(1)})$ which might consist of millions of parameters. $\w_t^{(1)}$ denotes the weights on the top layer combining the features $\phi(\x;\w_t^{(0)})$ and $\w_t^{(0)}$ carries all the internal representations.
Without loss of generality, we assume a sigmoid function for the discriminative probability
\begin{equation}
   q_t(y|\x; \W_t) = 1/(1+\exp\{-y<\w_t^{(1)}, \phi(\x;\w_t^{(0)})>\}).
\label{eq:qt}
\end{equation}
Both $\w_t^{(1)}$ and $\w_t^{(0)}$ can be learned by the standard stochastic gradient descent algorithm via backpropagation to minimize a cross-entropy loss with an additional term on the pseudo-negatives:
{\footnotesize
\begin{equation}
  \mathcal{L}(\W_t) = - \sum_{(\x_i,+1) \in S_+}^{i=1..n} \ln q_t(+1|\x_i; \W_t) - \sum_{(\x_i,-1) \in S_{-}^t}^{i=1..l} \ln q_t(-1|\x_i; \W_t)\nonumber \\ 
\label{eq:loss_binary}
\vspace{-5mm}
\end{equation}
}

\vspace{-2mm}
\subsubsection{Synthesis-step} \label{Synthesis}
\vspace{-2mm}
In the classification step, we obtain $q_t(y|\x; \W_t)$ which is then used to update
$p_t^-(\x)$ according to Eq. (\ref{eq:ICL}): 
\begin{equation}
\vspace{-2mm}
p_t^-(\x) =  \prod_{a=1}^t \frac{1}{Z_a} \frac{q_a (y=+1|\x; \W_a)}{q_a(y=-1|\x; \W_a)} p_{0}^-(\x).
\vspace{-0mm}
\label{eq:syn}
\end{equation}
In the synthesis-step, our goal is to draw fair samples from $p_t^-(\x)$.
The sampling process is carried out by backpropagation, but now we need to go through a sequence classifiers by using $\frac{1}{Z_a} \frac{q_a (y=+1|\x; \W_a)}{q_a(y=-1|\x; \W_a)}, a=1..t$. This can be time-consuming. In practice, we can simply perform backpropagation on the previous set $S_-^{t-1}$ by taking $\frac{1}{Z_t} \frac{q_t (y=+1|\x; \W_t)}{q_t(y=-1|\x; \W_t)}$. Therefore, generating pseudo-negative samples when training IGM does not need a large overhead.
Additional Gaussian noise can be added to the stochastic gradient as in \cite{welling2011bayesian} but we did not observe a big difference in the quality of samples in practice. This is probably due to the equivalent class \cite{wu2000equivalence} where the probability mass is widely distributed over an extremely large image space.

\noindent {\bf Sampling strategies}

In \cite{tu2007learning}, various Markov chain Monte Carlo techniques \cite{Jliu} including Gibbs sampling and Iterated Conditional Modes (ICM) have been adopted, which are often slow. Motivated by the DeepDream code \cite{mordvintsev2016deepdream} and Neural Artistic Style work \cite{gatys2015neural}, we perform stochastic gradient descent via backpropagation in synthesis.
Recent works show the connection and equivalence between stochastic gradient descent/ascent and Markov chain Monte Carlo sampling \cite{welling2011bayesian,chen2014stochastic,mandt2017stochastic}.
When conducting experiments, some alternative sampling schemes using SGD can be applied: i) early-stopping once $\x$ becomes positive (or after a small fixed number of steps); or ii) sampling with equilibrium after long steps. We found early-stopping effective and efficient, which can be viewed as contrastive divergence \cite{carreira2005contrastive} where a short Markov chain is simulated.

Note that the partition function (normalization) $Z_a$ is a constant that is not dependent on the sample $\x$. Let
\begin{equation}
g_t(\x) = \frac{q_t (y=+1|\x; \W_t)}{q_t(y=-1|\x; \W_t)}=\exp\{<\w_t^{(1)}, \phi(\x;\w_t^{(0)})>\},
\label{eq:logit}
\end{equation}
and take its $\ln$, which is nicely turned into the logit of $q_t(y=+1|\x; \W_t)$
\begin{equation}
\vspace{-2mm}
{\small
\ln g_t(\x) = <\w_t^{(1)},  \phi(\x;\w_t^{(0)})>.
}
\vspace{-0mm}
\label{eq:gt}
\end{equation}
Starting from a $\x$ drawn from $p_{t-1}^-(\x)$, we directly increase $<\w_t^{(1)}, \phi(\x;\w_t^{(0)})>$ using stochastic gradient ascent on $\x$ via backpropagation which allows us to
obtain fair samples subject to Eq. (\ref{eq:syn}). A noise can be injected as in \cite{welling2011bayesian} when performing SGD sampling.

\noindent {\bf Overall model}
The overall IGM model after $T$ stages of training becomes:
\begin{eqnarray}
\vspace{-2mm}
p_T^-(\x) &=& \frac{1}{Z} \prod_{t=1}^T  \frac{q_t (y=+1|\x; \W_t)}{q_t(y=-1|\x; \W_t)} p_{0}^-(\x) \nonumber \\
&=&  \frac{1}{Z} \prod_{t=1}^T \exp\{<\w_t^{(1)}, \phi(\x;\w_t^{(0)})>\} p_{0}^-(\x), \nonumber \\
\vspace{-2mm}
\label{eq:overall}
\end{eqnarray}
where $Z=\int \prod_{t=1}^T \exp\{<\w_t^{(1)}, \phi(\x;\w_t^{(0)})>\} p_{0}^-(\x) d\x$. 

IGM shares a similar cascade aspect with GDL \cite{tu2007learning} where the convergence of this iterative learning process to the target distribution was shown by the following theorem in \cite{tu2007learning}.
{\theor
\vspace{-1mm}
$KL[p(\x|y=+1)||p^-_{t+1}(\x)] \le KL[p(\x|y=+1)||p^-_{t}(\x)]$ where
$KL$ denotes the Kullback-Leibler divergences, and
$p(x|y=+1)\equiv p^+(x)$.
}

\subsection{An alternative: IGM-single}

\begin{figure}[!htb]
\vspace{-3mm}
\begin{center}
\begin{tabular} {c}
\hspace{-5mm} \includegraphics[width=0.4\textwidth]{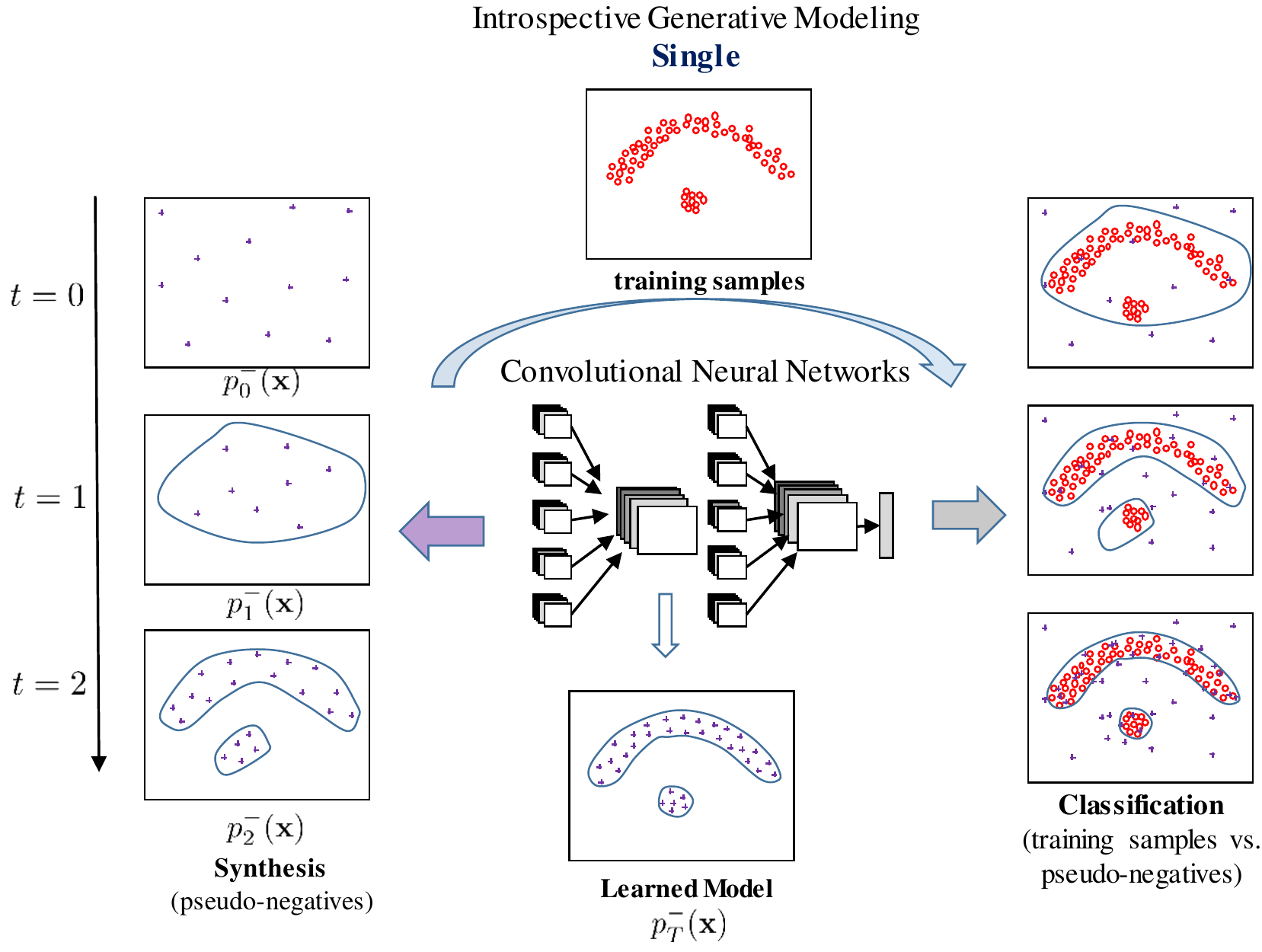} \\
\end{tabular}
\end{center}
\vspace{-3mm}
\caption{\footnotesize Schematic illustration of the pipeline of IGM-single.}
\label{fig:IGM_single}
\vspace{-2mm}
\end{figure}

We briefly present the IGM-single algorithm, which is similar to the introspective classifier learning algorithm \cite{jing2017Introspective} with the difference without the presence of input negative samples. The pipeline of IGM-single is shown in Fig. \ref{fig:IGM_single}. A key aspect here is that we maintain a single CNN classifier throughout the entire learning process. 

In the classification step, we obtain $q_t(y|\x; \W_t)$ (similar as Eq. \ref{eq:qt}) which is then used to update
$p_t^-(\x)$ according to Eq. (\ref{eq:IGM-single}): 
\begin{equation}
\vspace{-2mm}
p_t^-(\x) =  \frac{1}{Z_t} \frac{q_t (y=+1|\x; \W_t)}{q_t(y=-1|\x; \W_t)} p_{0}^-(\x).
\vspace{-0mm}
\label{eq:IGM-single}
\end{equation}
In the synthesis-step, we draw samples from $p_t^-(\x)$.

\noindent {\bf Overall model}
The overall IGM-single model after $T$ stages of training becomes:
\begin{equation}
\vspace{-0mm}
p_T^-(\x) = \frac{1}{Z_T} \exp\{<\w_T^{(1)}, \phi(\x;\w_T^{(0)})>\}  p_{0}^-(\x),
\vspace{-2mm}
\label{eq:overall}
\vspace{-1mm}
\end{equation}
where $Z_T= \int \exp\{<\w_T^{(1)}, \phi(\x;\w_T^{(0)})>\}  p_{0}^-(\x) d\x$.

\vspace{-2mm}
\subsection{Model-based anysize-image-generation}
\vspace{-2mm}
\begin{figure}[!htp]
\begin{center}
\begin{tabular} {c}
\includegraphics[width=0.4\textwidth]{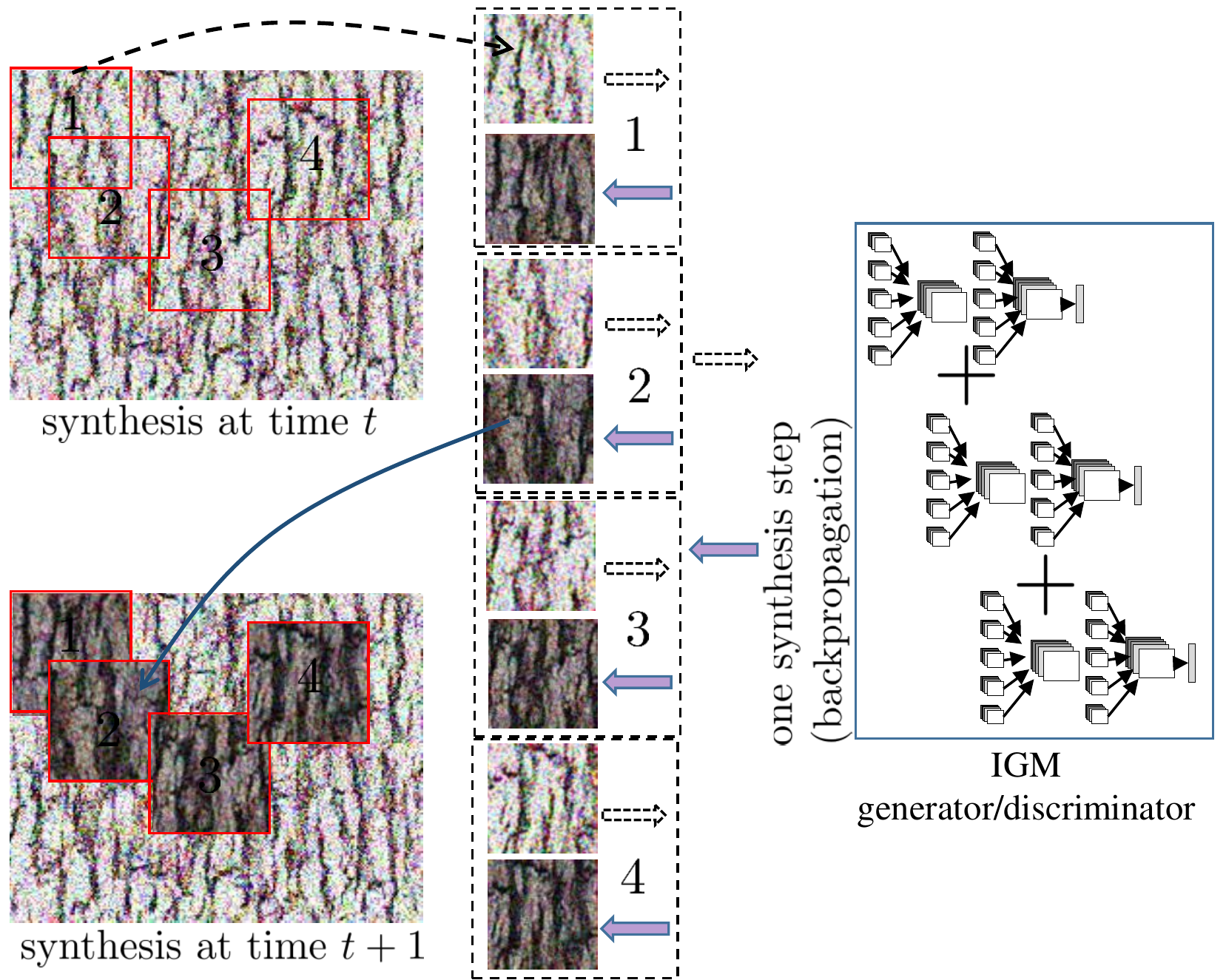}
\end{tabular}
\end{center}
\vspace{-3mm}
\caption{\footnotesize Illustration of model-based anysize-image-generation strategy.}
\label{fig:any}
\vspace{-4mm}
\end{figure}

Given a particularly sized image, anysize-image-generation within IGM allows one to generate/synthesize
an image much larger than the given one. Patches extracted from the training images are used in the training of the discriminator. However, their position within the training (or pseudo-negative) image is not lost. In particular, when performing synthesis using backpropagation, updates to the pixel values are made by considering the average loss of all patches that overlap a given pixel. Thus, up to round $T$, in order to consider the updates to the patch of $\x(i,j)$ centered at position $(i, j)$ for image ${\bf I}$ of size $m_1 \times m_2$, we perform backpropagation on the patches to increase the probability:
\begin{equation}
\vspace{-2mm}
p_T({\bf I}) \propto \prod_{t=1}^T \prod_{i=1}^{m_1} \prod_{j=1}^{m_2} g_t(\x(i,j)) p_{0}^-(\x(i,j))
\vspace{-0mm}
\label{eq:anysize-prob}
\end{equation}
where $g_t(\x(i,j))$ (see Eq. \ref{eq:logit}) denotes the score of the patch of size e.g. $64 \times 64$ for $\x(i,j)$ under the discriminator at round $t$. Fig. \ref{fig:any} gives an illustration for one round of sampling. This allows us to synthesize much larger images by being able to enforce the coherence and interactions surrounding a particular pixel. In practice, we add stochasticity and efficiency to the synthesis process by randomly sampling these set of patches.

\vspace{-2mm}
\section{Experiments}
\vspace{-2mm}

\begin{figure}[!htp]
\begin{center}
\begin{tabular} {c}
\hspace{-3mm} \includegraphics[width=0.45\textwidth]{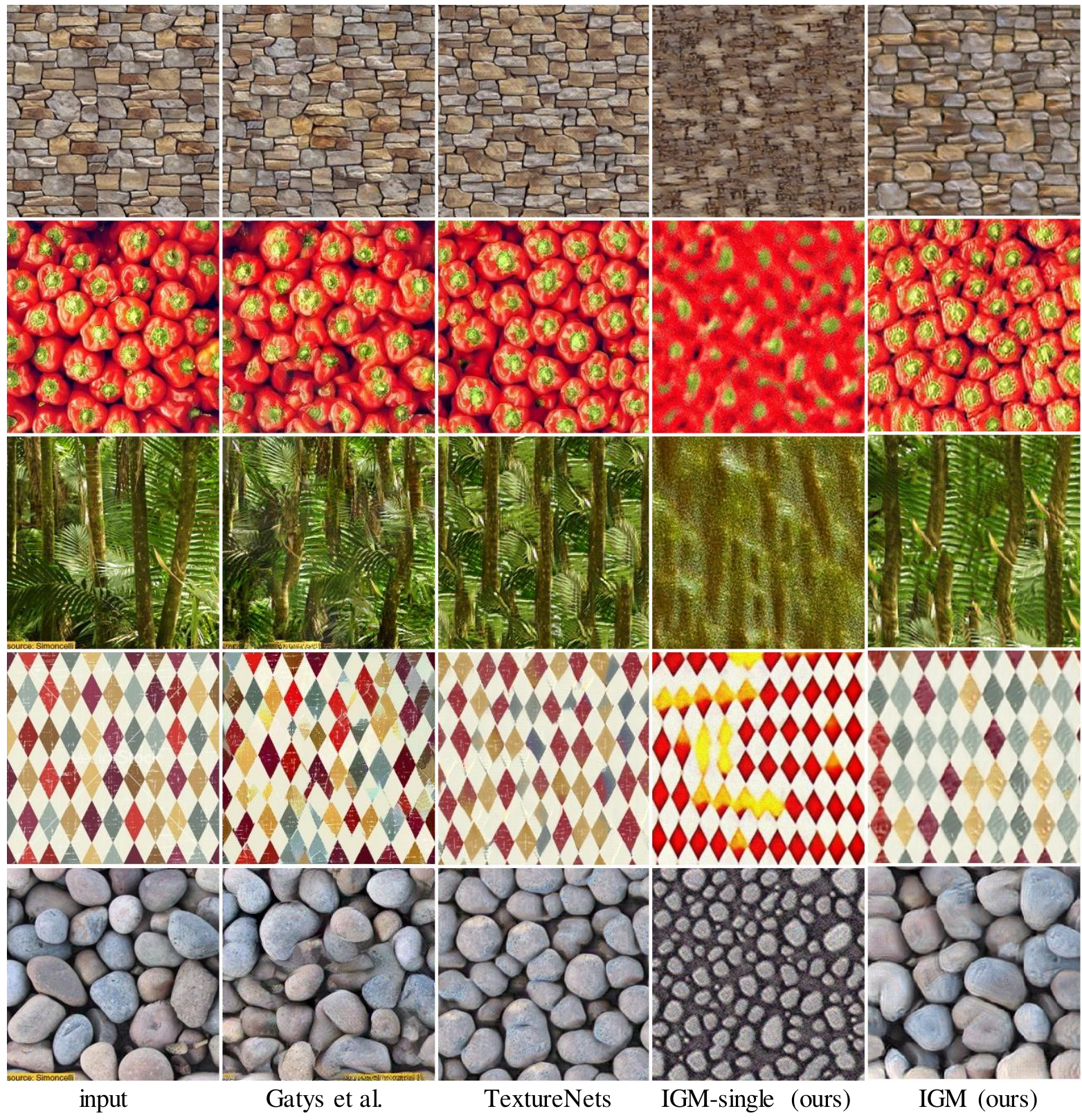}
\end{tabular}
\end{center}
\vspace{-3mm}
\caption{\footnotesize More texture synthesis results. Gatys et al. \cite{gatys2015neural} and Texture Nets \cite{ulyanov2016texture} results are from \cite{ulyanov2016texture}.}
\label{fig:texture_small}
\vspace{-2mm}
\end{figure}

We evaluate both IGM and IGM-single. In each method, we adopt the discriminator architecture of \cite{radford2015unsupervised}
which involves an input size of 64x64x3 in the RGB colorspace, four convolutional layers using $5 \times 5$ kernel sizes with the layers using  $64$, $128$, $256$ and $512$ channels, respectively. We include batch normalization after each convolutional layer (excluding the first) and use leaky ReLU activations with leak slope $0.2$. The classification layer flattens the input and finally feeds it into a sigmoid activation.

This serves as the discriminator for the $64 \times 64$ patches we extract from the training image(s). Note that is is a general purpose architecture with no modifications made for a specific task in mind.

In texture synthesis and artistic style, we make use of the ``anysize-image-generation'' architecture by adding a ``head'' to the network that, at each forward pass of the network, randomly selects some number (equal to the desired batch size) of $64 \times 64$ random patches (possibly overlapping) from the full sized images and passes them to the discriminator. This allows us to retain the whole space of patches within a training image rather than select some subset of them in advance to use during training.

\vspace{-2mm}
\subsection{Texture synthesis}
\vspace{-2mm}

Texture modeling/rendering is a long standing problem in computer vision and graphics \cite{heeger1995pyramid,zhu1997minimax,efros1999texture,portilla2000parametric}. Here we are interested in statistical texture modeling \cite{zhu1997minimax,xie2016cooperative}, instead of just texture rendering \cite{efros1999texture}.  
We train similar textures to \cite{ulyanov2016texture}. Each source texture is resized to $256 \times 256$, used as the single ``positive" example in the training set and a set of 200 negative examples are initially sampled from a normal distribution with $\sigma = 0.3$ of size $320 \times 320$ after adding padding of $32$ pixels to each spatial dimension of the image to ensure each pixel of the $256 \times 256$ center has equal probability of being extracted in some patch. 1000 patches are extracted randomly across the training images and fed to the discriminator at each forward pass of the network (during training and synthesis stages) from a batch size of 100 images --- 50 random positives and negatives when training and 100 pseudo-negatives during synthesis. At each round, our classifier is finetuned using stochastic gradient descent with learning rate $0.01$ from the previous round's classifier after the augmentation of the negative set with the 200 $320 \times 320$ synthesized pseudo-negatives. Pseudo-negatives from more recent rounds are chosen in mini-batches with higher probability than those of earlier rounds in order to ensure the discriminator learns from its most recent mistakes as well as provide for more efficient training when the set of accumulated negatives has grown large in later rounds. During the synthesis stage, pseudo-negatives are synthesized using the previous round's pseudo-negatives as their initialization. Adam is used with a learning rate of $0.1$ and $\beta = 0.5$ and stops early when the average probability of the patches under the discriminator is more likely than not to be a positive across some window of previous steps, usually 20, in order to reduce variance. This allows us to, on average, cross the decision boundary of the current iteration of the discriminator. We find this sampling strategy to attain a good balance in effectiveness and efficiency. Empirically, we find training the networks for 70 rounds to provide good results in terms of synthesis and distillation of the model's knowledge.

New textures are synthesized under IGM by: sampling from the same distribution used initially during training (in our case, normally distributed with $\sigma = 0.3$), performing backpropagation of the synthesis using the saved parameters of the networks for each round, and feeding the resulting partial synthesis to the next round. The same early stopping criterion is used as outlined during training, however, the number of patches is dialed down to match the number being synthesized. We use about 10 patches per image when synthesizing a $256 \times 256$ image since this matches the average number of patches extracted per image during training. Making the number of patches much larger than corresponding ratio used in the training process has shown to generate images of lower quality and diversity.

Under IGM-single there is a single network, and thus only a single round of synthesis takes place to transform the initial noise to a high probability texture.

Considering the results in Fig. \ref{fig:texture_big} and \ref{fig:texture_small}, we see that IGM generates images of similar quality to \cite{ulyanov2016texture}, however, it is usually more faithful to the structure of the input images. In Fig. \ref{fig:texture_big}, the ``bricks" texture synthesized by IGM is very strict about the grout lines being straight to ensure the bricks are rectilinear. Similarly, in Fig. \ref{fig:texture_small}, the ``forest" texture preserves continuity but allows for some of the variation in angle and path that the tree trunks take. The ``diamond" texture is reflective of the grid-like pattern seen from the input image and does not allow for overlap or differently sized diamonds. In the bottom row of ``pebbles", the resulting synthesis captures the size of the pebbles seen in the input image as well as the variation in color and shading.

\begin{figure}[!htp]
\begin{center}
\begin{tabular} {c}
\hspace{-2mm} \includegraphics[width=0.45\textwidth]{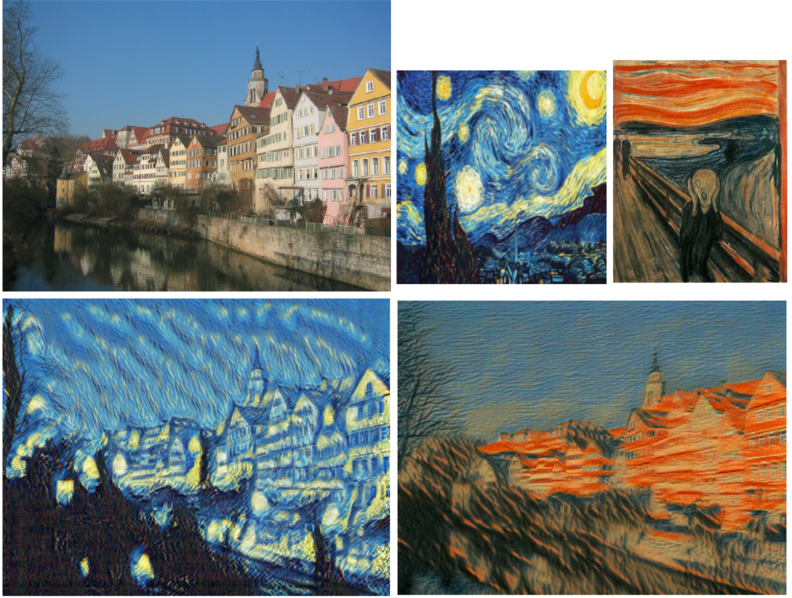} \\
\end{tabular}
\end{center}
\vspace{-3mm}
\caption{\footnotesize Artistic style transfer results using the ``Starry Night'' and ``Scream'' style on the image from Amsterdam.}
\label{fig:artistic_transfer}
\vspace{-3mm}
\end{figure}

\vspace{-2mm}
\subsection{Artistic style transfer}
\vspace{-2mm}

We also attempt to transfer artistic style as shown in \cite{gatys2015neural}. However, our architecture makes no use of additional networks for content and texture transferring task uses a loss functions during synthesis to minimize \[
-\ln p({\bf I}_{style} \mid {\bf I}) \propto \alpha \cdot ||{\bf I}_{style}-{\bf I}||_2 - (1-\alpha) \cdot\ln p_{style}^-({\bf I}_{style}),
\]
where ${\bf I}$ is an input image and ${\bf I}_{style}$ is its stylized version, and $p_{style}^-({\bf I})$ denotes the model learned from the training style image.
We include a  $L_2$ fidelity term during synthesis, weighted by a parameter $\alpha$, making ${\bf I}_{style}$ not too far away from the input image ${\bf I}$. We choose $\alpha = 0.3$ and average the $L_2$ difference between the original content image and the current stylized image at each step of synthesis. Two examples of the artistic style transfer are shown in Fig. \ref{fig:artistic_transfer}.

\vspace{-2mm}
\subsection{Face modeling}
\vspace{-2mm}

\begin{figure}[!htb]
\vspace{-1mm}
\begin{center}
\begin{tabular} {c}
\hspace{-5mm} \includegraphics[width=0.42\textwidth]{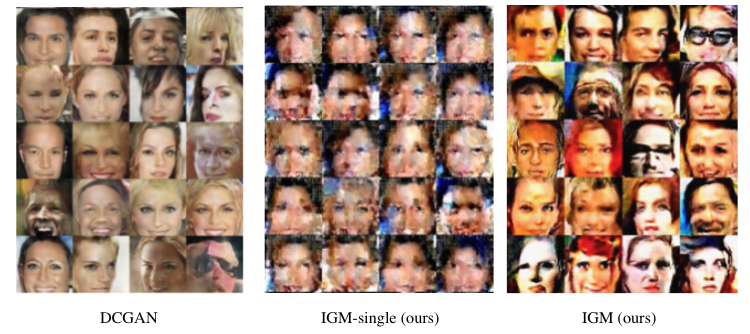}
\end{tabular}
\end{center}
\vspace{-5mm}
\caption{\footnotesize Generated images learned on the CelebA dataset. The first, the second, and the third column are respectively results by DCGAN \cite{radford2015unsupervised} using tensorflow implementation \cite{dcgan-tensorflow}, IGM-single, and IGM. 
}
\label{fig:face-compare}
\vspace{-2mm}
\end{figure}

The CelebA dataset \cite{liu2015faceattributes} is used in our face modeling experiment, which consists of $202,599$ face images. We crop the center $64 \times 64$ patches in these images as our positive examples. For the classification step, we use stochastic gradient descent with learning rate $0.01$ and a batch size of $100$ images, which contains $50$ random positives and $50$ random negatives. For the synthesis step, we use the Adam optimizer with learning rate $0.01$ and $\beta=0.5$ and stop early when the pseudo-negatives cross the decision boundary. In Fig. \ref{fig:face-compare}, we show some face examples generated by our model and the DCGAN model.

\vspace{-2mm}
\subsection{SVHN unsupervised learning} \label{sec:svhn_unsupervised}
\vspace{-2mm}

\begin{figure}[!htb]
\vspace{-1mm}
\begin{center}
\begin{tabular} {c}
\hspace{-5mm} \includegraphics[width=0.42\textwidth]{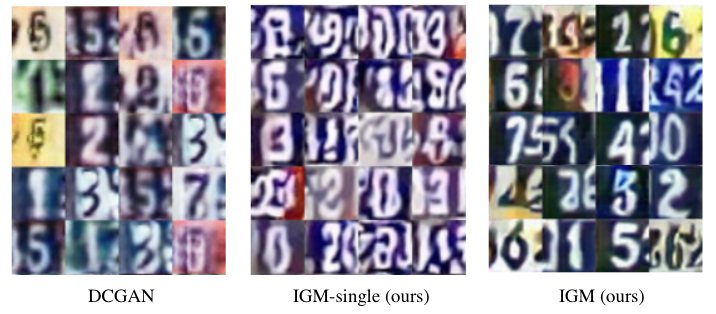}
\end{tabular}
\end{center}
\vspace{-5mm}
\caption{\footnotesize Generated images learned on the SVHN dataset.
The first, the second, and the third column are respectively results by DCGAN \cite{radford2015unsupervised} using tensorflow implementation \cite{dcgan-tensorflow}, IGM-single, and IGM.}
\label{fig:svhn-compare}
\vspace{-2mm}
\end{figure}

The SVHN \cite{netzer2011reading} dataset consists of color images of house numbers collected by Google Street View. The training set consists of $73,257$ images, the extra set consists of $531,131$ images, and the test set has $26,032$ images. The images are of the size $32 \times 32$. We combine the training and extra set as our positive examples for unsupervised learning. Following the same settings in the face modeling experiments, our IGM model can generate examples as shown in Fig. \ref{fig:svhn-compare}. 

\vspace{-2mm}
\subsection{SVHN semi-supervised classification}
\vspace{-2mm}

We perform the semi-supervised classification experiment by following the procedure outlined in \cite{radford2015unsupervised}. 
We first train a model on the SVHN training and extra set in an unsupervised way, as in Section \ref{sec:svhn_unsupervised}. Then, we train an L2-SVM on the learned representations of this model.
The features from the last three convolutional layers are concatenated to form a $14336$-dimensional feature vector. A $10,000$ example held-out validation set is taken from the training set and is used for model selection. The SVM classifier is trained on $1000$ examples taken at random from the remainder of the training set. The test error rate is averaged over $100$ different SVMs trained on random $1000$-example training sets. Within the same setting, our IGM model achieves the test error rate of $36.44 \pm 0.72\%$ and the DCGAN model achieves $33.13 \pm 0.83\%$ (we ran the DCGAN code \cite{dcgan-tensorflow} in an identical setting as IGM for a fair comparison since the result reported in \cite{radford2015unsupervised} was achieved by training on the ImageNet dataset).

\vspace{-2mm}
\section{Conclusion}
\vspace{-2mm}
Introspective generative modeling points to an encouraging direction for unsupervised image modeling that capitalizes on the power of discriminative deep convolutional neural networks. It can be adopted for a wide range of problems in computer vision and machine learning.

\section{Acknowledgement}
This work is supported by NSF IIS-1618477
and a Northrop Grumman Contextual Robotics grant. We thank Saining Xie, Jun-Yan Zhu, Jiajun Wu, Stella Yu, and Alexei Efros for helpful discussions.

{\small
\bibliographystyle{ieee}
\bibliography{egbib}
}

\end{document}